\RequirePackage{fix-cm}
\documentclass[twocolumn]{svjour3}          % twocolumn
\smartqed  % flush right qed marks, e.g. at end of proof

\usepackage{epsfig}
\usepackage{times}
\usepackage{url}            % simple URL typesetting
\usepackage{booktabs}       % professional-quality tables
\usepackage{amsfonts}       % blackboard math symbols
\usepackage{nicefrac}       % compact symbols for 1/2, etc.
\usepackage{microtype}      % microtypography
\usepackage{xcolor}         % colors
\usepackage{xspace}
\usepackage{graphicx}
\usepackage{wrapfig}
\usepackage{amsmath}
\usepackage{amssymb}
\usepackage{bm}
\usepackage{tabularx}
\usepackage{paralist}
\usepackage{natbib}
\usepackage{bbding}
\usepackage[linesnumbered,ruled,vlined]{algorithm2e}

\newcommand{\Tref}[1]{Table~\ref{#1}}
\newcommand{\Eref}[1]{Equation~(\ref{#1})}

\newcommand{\Sref}[1]{Section~\ref{#1}}

\newcommand{\fref}[1]{Fig.~\ref{#1}}

\makeatletter
\DeclareRobustCommand\onedot{\futurelet\@let@token\@onedot}
\def\@onedot{\ifx\@let@token.\else.\null\fi\xspace}
\def\eg{\emph{e.g}\onedot} 
\def\ie{\emph{i.e}\onedot} 
 
\def\etc{\emph{etc}\onedot} \def\vs{\emph{vs}\onedot}

\makeatother
\begin{document}

\title{Learning to Deblur Polarized Images}
% \subtitle{Do you have a subtitle?\\ If so, write it here}


\author{Chu Zhou$^1\textsuperscript{\textdagger}$ \and Minggui Teng$^{2,3}$ \and Xinyu Zhou$^4$ \and Chao Xu$^4$ \and Imari Sato$^1$ \and \\Boxin Shi$^{2,3}$}

\authorrunning{Chu Zhou et al.} % if too long for running head

\institute{
\Envelope \, Boxin Shi \at
\email{shiboxin@pku.edu.cn} \vspace{0.1cm}
\and
\textdagger \, Most of this work was done as a PhD student at Peking University.\\ 
$1$ \, National Institute of Informatics, Japan\\
$2$ \, National Engineering Research Center of Visual Technology, School of Computer Science, Peking University, China\\
$3$ \, State Key Laboratory for Multimedia Information Processing, School of Computer Science, Peking University, China\\
$4$ \, National Key Laboratory of General AI, School of Intelligence Science and Technology, Peking University, China
}

\date{Received: date / Accepted: date}

\maketitle

\begin{abstract}
A polarization camera can capture four linear polarized images with different polarizer angles in a single shot, which is useful in polarization-based vision applications since the degree of linear polarization (DoLP) and the angle of linear polarization (AoLP) can be directly computed from the captured polarized images. However, since the on-chip micro-polarizers block part of the light so that the sensor often requires a longer exposure time, the captured polarized images are prone to motion blur caused by camera shakes, leading to noticeable degradation in the computed DoLP and AoLP. Deblurring methods for conventional images often show degraded performance when handling the polarized images since they only focus on deblurring without considering the polarization constraints. In this paper, we propose a polarized image deblurring pipeline to solve the problem in a polarization-aware manner by adopting a divide-and-conquer strategy to explicitly decompose the problem into two less ill-posed sub-problems, and design a two-stage neural network to handle the two sub-problems respectively. Experimental results show that our method achieves state-of-the-art performance on both synthetic and real-world images, and can improve the performance of polarization-based vision applications such as image dehazing and reflection removal.

\keywords{Image deblurring \and Polarized images \and Polarization camera \and Deep learning}
\end{abstract}

\section{Introduction}
    Polarization is an important property of light in addition to its amplitude and phase. By virtue of the physical information encoded in the polarization-relevant parameters (\eg, the degree of linear polarization (DoLP) and the angle of linear polarization (AoLP) of the incoming light to the sensor), it has demonstrated unique advantages in various computer vision tasks, such as glass segmentation \citep{mei2022glass}, color constancy \citep{ono2022degree}, and shape reconstruction \citep{deschaintre2021deep}. Besides, it also plays a crucial role in applications for autonomous driving, such as road detection \citep{li2020full}, reflection removal \citep{lyu2019reflection, lyu2022physics}, dehazing \citep{zhou2021learning}, and HDR reconstruction \citep{zhou2023polarizationHDR}, conquering the bottleneck of conventional vision algorithms in this field.
    
    With the development of the division of focal plane (DoFP) technology, modern polarization cameras (\eg, Lucid Vision Phoenix polarization camera\footnote{\label{footnote: Lucid}https://thinklucid.com/product/phoenix-5-0-mp-polarized-model/}) can capture four spatially-aligned and temporally-synchronized linear polarized images with different polarizer angles in a snapshot, bringing convenience to the acquisition of the DoLP and AoLP. However, the captured polarized images are prone to motion blur caused by camera shakes, since the on-chip micro-polarizers block part of the light so that the sensor often requires a longer exposure time. Therefore, it is of practical interest to deblur polarized images for acquiring the DoLP and AoLP accurately.
    
    Recent advances in image deblurring \citep{nah2017deep, kupyn2019deblurgan, cho2021rethinking, ji2022xydeblur} have shown their effectiveness in a large variety of scenes. They adopt deep neural networks to remove the motion blur in a single unpolarized image by extracting image features and priors from a large amount of training data. However, deblurring multiple polarized images (captured by a polarization camera in a snapshot) could be a much more challenging problem than deblurring a single unpolarized image due to several fundamental issues:
    \begin{itemize}
        \item[$\bullet$] Polarization constraints (the numerical relationship between multiple polarized images constrained by the physical meanings of polarization) have to be preserved across multiple images in addition to removing the motion blur, making the problem highly ill-posed. Since the deblurring methods designed for conventional images can only process the polarized images in a polarization-unaware manner (\ie, treat multiple polarized images as ordinary frames and ignore the physical meanings of polarization), they have no chance to take the preservation of polarization constraints into consideration.
        \item[$\bullet$] The DoFP technology adopted by modern polarization cameras induces a higher level of sensor noise (compared to the same sensor without micro-polarizers) \citep{kumar2007motion, qiu2019polarization, abubakar2018block}, leading to degradation of low-level features (\eg, structural and boundary information). Since current deblurring methods usually aim to deal with the images captured by conventional sensors, they cannot take advantage of the information and priors provided by the polarization-related parameters to mitigate the difficulties during the feature extraction and fusion procedure.
        \item[$\bullet$] Since the on-chip micro-polarizers attenuate scene radiance and the extent of attenuation is scene-dependent \citep{wu2020hdr, zhou2023polarizationHDR}, the exposures turn out to be spatially-variant and arbitrary, making high-level features (\eg, semantic and contextual information) less distinctive. Since recent deblurring methods usually adopt a single-stage pipeline that depends strongly on the quality of high-level features to hallucinate the sharp contents to restore the sharp image from its blurry counterpart in an end-to-end manner, they tend to suffer from ringing artifacts and fake textures.
    \end{itemize}
    
    In this paper, we propose a polarized image deblurring pipeline to solve the problem in a polarization-aware manner, by adopting a divide-and-conquer strategy to explicitly decompose the problem into two less ill-posed sub-problems: deblurring the corresponding unpolarized image for restoring the sharp image content, without considering the polarization constraints; utilizing the restored sharp image content to guide the deblurring procedure of the polarized images, considering the recovery of the polarization constraints. Concretely, we have two key observations on the characteristics of polarized images: 
    \begin{itemize}
        \item[(1)] \textit{Blur resistance property of Stokes parameters}: two (out of three) Stokes parameters (which are polarization-related parameters and will be introduced later) have similar appearance to the gradient distribution of the corresponding unpolarized image and are less affected by motion blur.
        \item[(2)] \textit{Semantic similarity between polarized and unpolarized images}: the polarized images usually have similar semantic and contextual information to the corresponding unpolarized image.
    \end{itemize}
     Based on these inspiring characteristics of polarized images, we specially build a two-stage neural network to handle the two sub-problems respectively: The first stage is an \textit{unpolarized image estimator} that leverages the less-contaminated structural and boundary information encoded in the Stokes parameters for repairing degraded low-level features. The second stage is a \textit{polarized image reconstructor} that transfers the semantic and contextual information from the estimated unpolarized image to the polarized images with robustly extracted high-level features. To summarize, this paper makes contributions by demonstrating three customized model designs to deal with the three issues mentioned above:
    \begin{itemize}
        \item[$\bullet$] a polarized image deblurring pipeline to solve the problem in a polarization-aware manner, by decomposing it into two less-ill-posed sub-problems and handling them via:
        \item[$\bullet$] leveraging the less-contaminated structural and boundary information encoded in the Stokes parameters for repairing degraded low-level features; and
        \item[$\bullet$] transferring semantic and contextual information from the estimated unpolarized image to the polarized images with robustly extracted high-level features.
    \end{itemize}
    
    Experimental results show that our method achieves state-of-the-art performance on both synthetic and real-world images, and can improve the performance of polarization-based vision applications such as image dehazing and reflection removal.

\section{Related work}
    \subsection{Deblurring unpolarized images}
    Generally, methods for deblurring unpolarized images can be divided into two categories: image deblurring and video deblurring methods. We will have an overview on them respectively in the following.
    
    \textbf{Image deblurring methods.} Previous works turned the problem into a maximum \textit{a posteriori} (MAP) estimation problem and adopted numerical optimization to enforce sharpness \citep{fergus2006removing, xu2013unnatural, pan2016l_0, yang2019variational, chen2020enhanced, chen2021blind, zhang2022pixel}. Recently, learning-based approaches have been introduced to deal with this problem \citep{zhang2022deep}. By extracting features from a large amount of training data to reduce the ill-posedness, they often show better performance and higher efficiency than numerical optimization-based methods. Some works proposed to restore the sharp image from its blurry counterpart in an end-to-end manner using neural networks with different kinds of architectures, including convolutional neural network (CNN) \citep{nah2017deep, zhang2018dynamic, tao2018scale, gao2019dynamic, zhang2019deep, suin2020spatially, cho2021rethinking, chi2021test, mou2022deep, ji2022xydeblur, chen2022simple, liu2024motion}, generative adversarial network (GAN) \citep{kupyn2018deblurgan, kupyn2019deblurgan, zhang2020deblurring, wen2021structure, li2022learning}, multi-layer perceptron (MLP) \citep{tu2022maxim}, diffusion probabilistic model (DPM) \citep{whang2022deblurring, ren2023multiscale}, vision transformer (ViT) \citep{zamir2022restormer, wang2022uformer, tsai2022stripformer, chu2022improving, kong2023efficient}, \etc, demonstrating visually impressive results. Some works proposed to deblur in a two-stage manner, by adopting neural networks to estimate blur-related physical parameters (\eg, blur kernels and motion flows) first and then utilizing them to guide the deblurring process \citep{kaufman2020deblurring, ren2020neural, dong2021learning, tran2021explore, chen2021learning, chakrabarti2016neural, sun2015learning, chen2018reblur2deblur, gong2017motion, yuan2020efficient}, showing good generalization ability. Some methods adopted unrolling algorithms to bridge the gap between classical MAP optimization and pure deep learning techniques, which are also useful in deblurring or other image enhancement tasks \citep{kruse2017learning, eboli2020end, zhang2020deep}. However, they are not suitable for handling multiple polarized images since they can only deblur in a polarization-unaware manner by processing one polarized image at a time, without considering the polarization constraints.
    
    \textbf{Video deblurring methods.} Video deblurring methods can process multiple frames instead of a single frame at a time. Early works adopted numerical optimization to perform blur kernel estimation and deconvolution for restoring sharp frames \citep{bar2007variational, wulff2014modeling, ren2017video}. With the development of deep-learning, neural networks have also been adopted to handle this problem. These methods could be divided into three categories: window-based methods, recurrent methods, and memory-based methods. Window-based methods take multiple frames as input and try to reconstruct the middle frame, by referencing neighboring frames within a small fixed-size window \citep{su2017deep, pan2020cascaded, suin2021gated, zhang2022spatio}. Recurrent methods deblur frames in sequence, by collectively aggregating frame features till up-to-date to provide restoration cues \citep{wieschollek2017learning, hyun2017online, nah2019recurrent, zhou2019spatio, zhong2020efficient, li2021arvo, zhu2022deep, jiang2022erdn, wang2022efficient}. Memory-based methods allow the feature propagation from the entire sequence when restoring the given frame, by using memory banks to provide appropriate spatio-temporal information for each region and exploiting the features of all frames \citep{ji2022multi}. Additionally, there are methods designed for deblurring burst images \citep{aittala2018burst, wieschollek2017end}, which can also process multiple frames. However, they are still not suitable for handling multiple polarized images since the polarized images do not contain temporal information as the video frames or burst images.

    \subsection{Polarization and image enhancement}
    Here, we will introduce the intersection of polarization and image enhancement in two aspects: enhancing polarized images and polarization-based image enhancement.
    
    \textbf{Enhancing polarized images.} There are many methods aiming to enhance the polarized images (\eg, joint chromatic and polarimetric demosaicing \citep{tyo2009total, qiu2019polarization, wen2019convolutional, wen2021sparse, pistellato2022deep, liu2022enhanced, yi2024demosaicking, guo2024attention}, denoising \citep{zhang2017pca, abubakar2018block, tibbs2018denoising, li2020learning, liang2022bm3d, li2023polarized}, and low-light enhancement \citep{hu2020iplnet, li2022polarized, xu2022colorpolarnet, zhou2023polarization}) to acquire the DoLP and AoLP accurately for improving the performance of downstream applications of polarization-based vision. Some methods perform enhancement in a direct manner, by processing multiple polarized images simultaneously in the intensity domain (\eg, IPLNet \citep{hu2020iplnet}). However, their performance may not that good, since enhancing in the intensity domain cannot make effective use of the information provided by polarization. On the contrary, some methods adopt a Stokes-domain enhancement pipeline (\eg, ColorPolarNet \citep{xu2022colorpolarnet} and PLIE \citep{zhou2023polarization}) to enhance the polarized images in an indirect manner. These methods usually show higher performance and robustness, since the polarization characteristics of the incoming light to the sensor are fully encoded in the Stokes parameters, making the enhancement process more polarization-aware. Although the above methods are specially designed for processing multiple polarized images, the enhancement tasks targeted by them have completely different properties from the deblurring problem.

    \textbf{Polarization-based image enhancement.} Different from enhancing polarized images, polarization-based image enhancement aims to utilize the polarization information to handle the image enhancement problems. Since the reflection and scattering phenomenons would alter the polarization property of light, polarization information can be used as priors to eliminate the reflection artifacts \citep{lyu2019reflection, lei2020polarized, lyu2022physics} or remove the haze \citep{zhou2021learning, ma2024multi, ma2024image} inside the images. Since the polarization cameras can not only offer spatially-variant exposures but also contain abundant structural and contextual information of the scene (encoded in the DoLP and AoLP), we can also use polarization information to achieve HDR reconstruction \citep{wu2020hdr, ting2021, zhou2023polarizationHDR}). Although the above methods may also utilize the polarization-related parameters to provide guidance, their underlying strategies and objectives diverge from those of methods designed for enhancing polarized images, as they are designed to solve different types of problems.

    \begin{figure*}[t]
	\centering
        \includegraphics[width=1.0\linewidth]{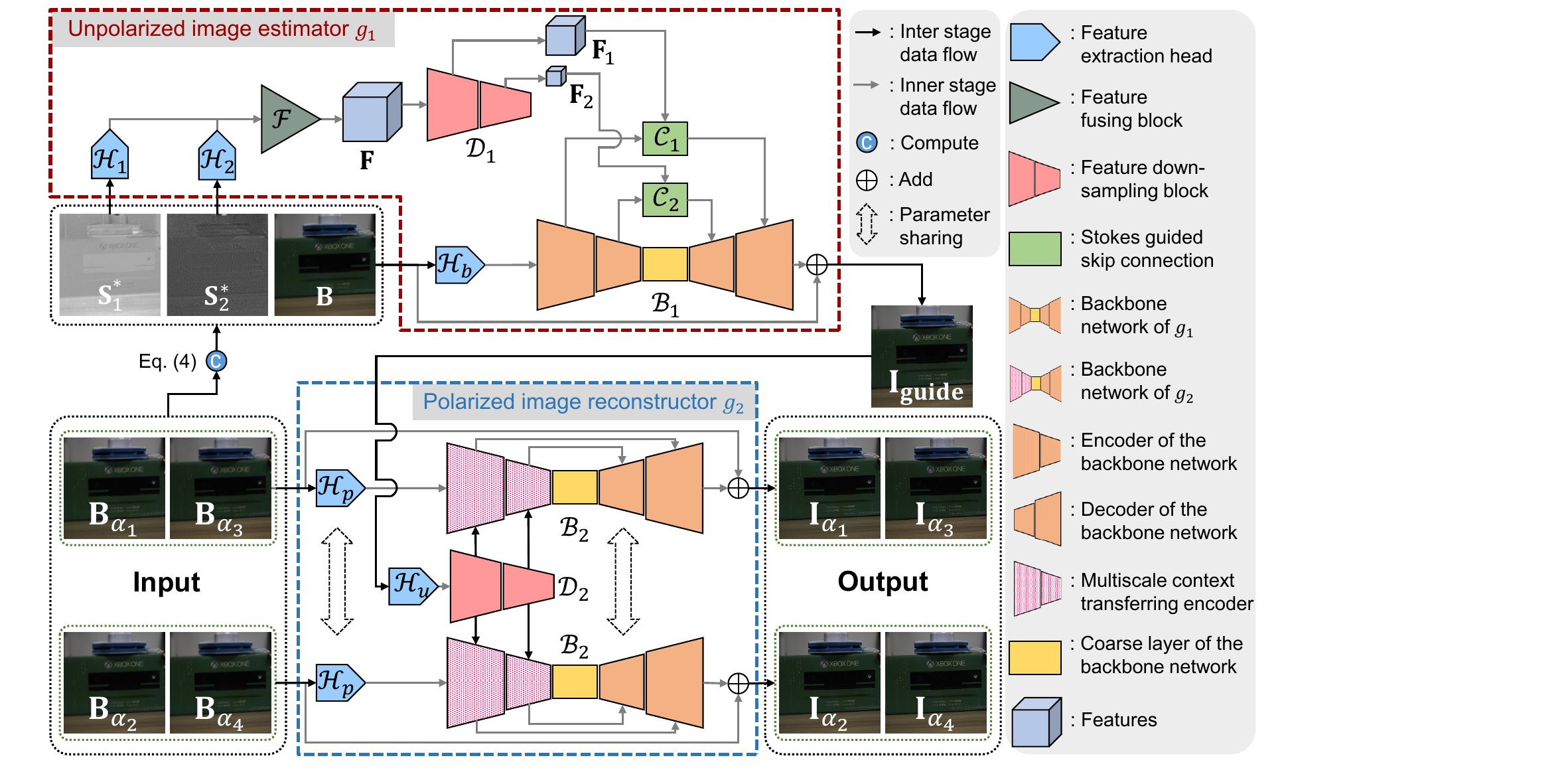}
        \caption{Architecture of the neural network that implements the proposed polarized image deblurring pipeline. It consists of two stages: an unpolarized image estimator that uses $\mathbf{S}^*_{1,2}$ (the blurry counterparts of the Stokes parameters $\mathbf{S}_{1,2}$) as priors to estimate the sharp unpolarized image $\mathbf{I}_{\text{guide}}$, and a polarized image reconstructor that reconstructs $\mathbf{I}_{\alpha_{1,2,3,4}}$ under the guidance of $\mathbf{I}_{\text{guide}}$. All images are normalized for visualization.}
        \label{fig: Network}
    \end{figure*}

\section{Polarized image deblurring pipeline}
    \label{sec: Pipeline}
    Denoting the total intensity of the incoming light to the sensor as $\mathbf{I}$ (\ie, the unpolarized image captured without polarizers), when placing a linear polarizer in front of the camera, the captured polarized image $\mathbf{I}_\alpha$ can be calculated using Malus’ law \citep{hecht2002optics}:
    \begin{equation}
        \label{eq: Malus}
        \mathbf{I}_\alpha = \frac{1}{2}\mathbf{I} \cdot (1 - \mathbf{p} \cdot \cos(2(\alpha-\bm{\theta}))),
    \end{equation}
    where $\alpha$ is the polarizer angle, $\mathbf{p} \in [0,1]$ and $\bm{\theta} \in [0, \pi]$ denote the DoLP and AoLP of the incoming light to the sensor respectively. Note that in this paper, we assume the camera response function to be linear since polarization cameras usually output images with a linear camera response function, and focus on linear polarization (\ie, do not consider circular polarization) since polarization cameras only equip linear polarizers.
    
    By virtue of four-directional, on-chip micro-polarizers provided by the DoFP technology, a modern polarization camera can capture four spatially-aligned and temporally-synchronized linear polarized images $\mathbf{I}_{\alpha_{1,2,3,4}}$ with different polarizer angles $\alpha_{1,2,3,4}=0^{\circ}, 45^{\circ}, 90^{\circ}, 135^{\circ}$ in a snapshot, which brings convenience to the acquisition of $\mathbf{p}$ and $\bm{\theta}$. This is because $\mathbf{p}$ can be calculated using
    \begin{equation}
        \label{eq: DoP}
        \mathbf{p}=\frac{\sqrt{\mathbf{S}_1^2 + \mathbf{S}_2^2}}{\mathbf{S}_0},
    \end{equation}
    and $\bm{\theta}$ can be calculated using
    \begin{equation}
        \label{eq: AoP}
        \bm{\theta}=\frac{1}{2}\arctan(\frac{\mathbf{S}_2}{\mathbf{S}_1}),
    \end{equation}
    where $\mathbf{S}_{0,1,2}$ are called the Stokes parameters \citep{konnen1985polarized} of the incoming light to the sensor, which can be directly obtained from the captured polarized images $\mathbf{I}_{\alpha_{1,2,3,4}}$ according to their physical meanings\footnote{$\mathbf{S}_0$ describes the total intensity of the light, and $\mathbf{S}_1$ ($\mathbf{S}_2$) describes the difference between the intensity of the vertical ($135^{\circ}$) and horizontal ($45^{\circ}$) polarized light \citep{konnen1985polarized}.}:
    \begin{equation}
        \label{eq: StokesParameters}
        \begin{cases}
        \mathbf{S}_0 = \mathbf{I} = \frac{1}{2}(\mathbf{I}_{\alpha_1} + \mathbf{I}_{\alpha_2} + \mathbf{I}_{\alpha_3} + \mathbf{I}_{\alpha_4})\\
        \mathbf{S}_1 = \mathbf{I}_{\alpha_3} - \mathbf{I}_{\alpha_1}\\
        \mathbf{S}_2 = \mathbf{I}_{\alpha_4} - \mathbf{I}_{\alpha_2}
        \end{cases}
        .
    \end{equation}
    When motion blur occurs, denoting the blurry counterparts of $\mathbf{I}$ and $\mathbf{I}_{\alpha_i}$ as $\mathbf{B}$ and $\mathbf{B}_{\alpha_i}$ respectively ($i=1,2,3,4$), their relationships can be written as
    \begin{equation}
        \label{eq: Blur1}
        \mathbf{B} = \mathbf{I} \otimes \mathbf{K} + \mathbf{N},
    \end{equation}
    and
    \begin{equation}
        \label{eq: Blur2}
        \mathbf{B}_{\alpha_i} = \mathbf{I}_{\alpha_i} \otimes \mathbf{K} + \mathbf{N}_i (i=1,2,3,4),
    \end{equation}
    where $\otimes$ stands for the convolution operation, $\mathbf{K}$ denotes a spatially-variant blur kernel determined by the motion field, $\mathbf{N}$ and $\mathbf{N}_i$ represent noise terms depending on the input signal and camera configuration \citep{hu2020iplnet, xu2022colorpolarnet, zhou2023polarization}. Note that both $\mathbf{B}$ and $\mathbf{B}_{\alpha_i}$ ($i=1,2,3,4$) can be considered to share the same blur kernel $\mathbf{K}$, since the pixel shifts among the polarized images caused by mosaicing is small enough (usually 1 pixel for a $4\times4$ mosaicing of polarization and color filters inside a polarization camera), which will not cause significant changes to the motion field and is often ignored by other works designed for enhancing the polarized images \citep{hu2020iplnet, li2022polarized, xu2022colorpolarnet, zhou2023polarization}.
    
    We aim to restore $\mathbf{p}$ and $\bm{\theta}$ from their degraded counterparts $\mathbf{p}^*$ and $\bm{\theta}^*$ by deblurring $\mathbf{B}_{\alpha_{1,2,3,4}}$. Unlike conventional image deblurring problems which only need to focus on removing the blurry artifacts, our problem becomes more ill-posed since we need to consider how to preserve the polarization constraints across the polarized images in addition to deblurring. This is because according to \Eref{eq: DoP}, \Eref{eq: AoP}, and \Eref{eq: StokesParameters}, the relationship among $\mathbf{p}$, $\bm{\theta}$, and $\mathbf{I}_{\alpha_{1,2,3,4}}$ involves non-linear operations, which makes the accuracy of $\mathbf{p}$ and $\bm{\theta}$ highly sensitive to the preservation of the polarization constraints \citep{hu2020iplnet, zhou2023polarization}. Therefore, we adopt a divide-and-conquer strategy to explicitly decompose the problem into two less ill-posed sub-problems, and propose a two-stage polarized image deblurring pipeline to handle them respectively in a polarization-aware manner: 
    \begin{itemize}
        \item[$\bullet$] The first stage aims to deblur the corresponding unpolarized image for restoring the sharp image content, without considering the polarization constraints.
        \item[$\bullet$] The second stage aims to utilize the restored sharp image content to guide the deblurring procedure of the polarized images, considering the recovery of the polarization constraints.
    \end{itemize}
    This pipeline is implemented by building a two-stage neural network, as shown in \fref{fig: Network}, and we will detail its model designs in the following section.
    
\section{Network model design}
    \subsection{Unpolarized image estimator}
    \label{subsec: Stage1}
\begin{figure*}[t]
	\centering
        \includegraphics[width=1.0\linewidth]{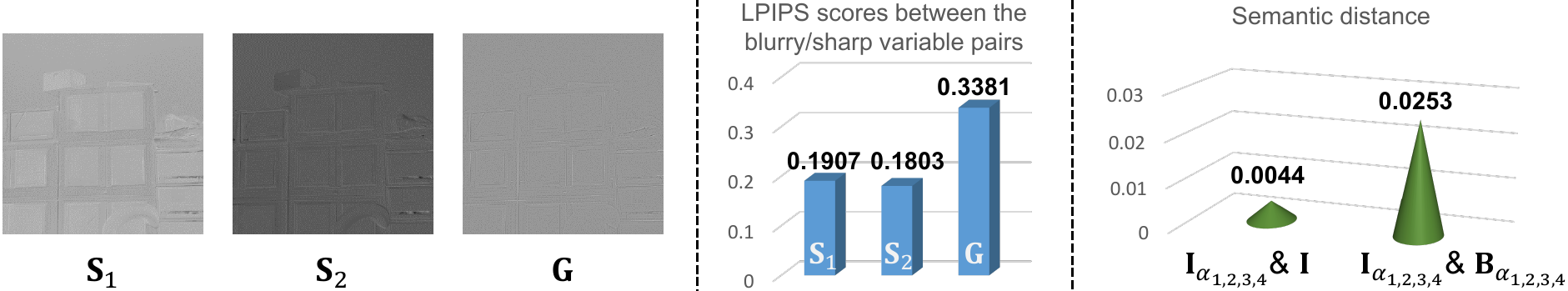}
        \caption{Left: Two of the Stokes parameters ($\mathbf{S}_{1,2}$) have similar ``appearance'' to the gradient distribution $\mathbf{G}$ of the corresponding unpolarized image (please zoom-in for better details). Middle: LPIPS \citep{zhang2018perceptual} scores between the blurry/sharp variable pairs, including $\mathbf{S}_{1,2}$ and $\mathbf{G}$. Right: Semantic distance between $\mathbf{I}_{\alpha_{1,2,3,4}}$ and $\mathbf{I}$, $\mathbf{I}_{\alpha_{1,2,3,4}}$ and $\mathbf{B}_{\alpha_{1,2,3,4}}$.}
        \label{fig: Verification}
    \end{figure*}

    \textbf{Overall architecture.} We design the first stage as an unpolarized image estimator $g_1$, aiming to estimate $\mathbf{I}$ from $\mathbf{B}$. Apparently, the problem handled by $g_1$ is similar to deblurring a single conventional image, however, since $\mathbf{B}$ is computed from $\mathbf{B}_{\alpha_{1,2,3,4}}$ which are captured by a polarization camera, the low-level features (\eg, structural and boundary information) would suffer from severe degradation, making the problem more challenging. This is because the DoFP technology adopted by modern polarization cameras induces higher level sensor noise (compared to the same sensor without micro-polarizers) \citep{kumar2007motion, qiu2019polarization, abubakar2018block} so that the noise term $\mathbf{N}$ in \Eref{eq: Blur1} and \Eref{eq: Blur2} becomes non-negligible. Prominently, we notice that there are two Stokes parameters ($\mathbf{S}_{1,2}$) having similar ``appearance'' to the gradient distribution $\mathbf{G}$ of the corresponding unpolarized image due to their ``differential'' nature (\ie, both of them describe the difference between two polarized images according to \Eref{eq: StokesParameters}), as shown in \fref{fig: Verification} (left); besides, $\mathbf{S}_{1,2}$ are less affected by motion blur since they often have relatively stronger response near the edges, which means that their blurry counterparts $\mathbf{S}^*_{1,2}$ could retain less-contaminated structural and boundary information. To verify it, we design a simulation experiment on our training set (including 8000 images, see \Sref{subsec: Dataset} for details) to quantitatively measure the gap between the sharp/blurry variable pairs, by computing the average value of LPIPS score (learned perceptual image patch similarity, which is suitable for measuring the similarity distance in the feature space) \citep{zhang2018perceptual}. Results are shown in \fref{fig: Verification} (middle), we can see that the degradation degree of $\mathbf{S}_{1,2}$ is nearly half of that of $\mathbf{G}$. Therefore, we expect that $\mathbf{S}^*_{1,2}$ can compensate the degraded low-level features, and decide to use them as priors. As shown in the red dashed frame of \fref{fig: Network}, this stage can be written as the following equation:
    \begin{equation}
        \label{eq: Stage1}
        \mathbf{I}_{\text{guide}} = g_1(\mathbf{B}, \mathbf{S}^*_{1,2}),
    \end{equation}
    where $\mathbf{I}_{\text{guide}}$ denotes the estimated sharp unpolarized image by $g_1$, which will serve as an important guiding signal in the next stage.

    \textbf{Layer details.} We first adopt two feature extraction heads (consisting of several convolution layers) $\mathcal{H}_1$ and $\mathcal{H}_2$ to extract features from $\mathbf{S}^*_1$ and $\mathbf{S}^*_2$ respectively, and adopt a feature fusing block (consisting of a convolution layer and a dense block \citep{huang2017densely}) $\mathcal{F}$ to obtain the fused Stokes features $\mathbf{F}$. Then, we adopt another feature extraction head $\mathcal{H}_b$ to extract features from $\mathbf{B}$, and feed them along with $\mathbf{F}$ into a backbone network $\mathcal{B}_1$ (a modified autoencoder architecture \citep{hinton2006reducing} similar to the one used in \citep{zhou2023polarization}) to learn the residual between $\mathbf{B}$ and $\mathbf{I}$. For fully utilizing the less-contaminated structural and boundary information retained in $\mathbf{S}^*_{1,2}$, we need to allow $\mathbf{F}$ to directly affect $\mathbf{B}$ across the entire feature domain. Therefore, instead of directly feeding it into the encoder of $\mathcal{B}_1$, we propose to use Stokes guided skip connections (SGSC) to substitute the skip connections inside $\mathcal{B}_1$ for processing it. SGSC accepts two kinds of inputs: the features from the backbone network and the features from the Stokes parameters. It uses residual bottleneck blocks \citep{he2016deep} to filter each feature channel, and uses a squeeze-and-excitation block \citep{hu2018squeeze} to recalibrate the features by learning normalized weights for recovering fine-grained high-order image structure. Specifically, we adopt a feature downsampling block (consisting of several convolution layers and residual blocks \citep{he2016deep}) $\mathcal{D}_1$ to downsample $\mathbf{F}$ to get multiscale features of the Stokes parameters ($\mathbf{F}_1$ and $\mathbf{F}_2$), and adopt two SGSCs ($\mathcal{C}_1$ and $\mathcal{C}_2$) to process them respectively.

    \subsection{Polarized image reconstructor}
    \label{subsec: Stage2}
    \textbf{Overall architecture.} We design the second stage as a polarized image reconstructor $g_2$, aiming to reconstruct $\mathbf{I}_{\alpha_{1,2,3,4}}$ from $\mathbf{B}_{\alpha_{1,2,3,4}}$ under the guidance of $\mathbf{I}_{\text{guide}}$. As shown in the blue dashed frame of \fref{fig: Network}, this stage can be written as the following equation:
    \begin{equation}
        \label{eq: Stage2}
        \mathbf{I}_{\alpha_{1,2,3,4}} = g_2(\mathbf{B}_{\alpha_{1,2,3,4}}, \mathbf{I}_{\text{guide}}).
    \end{equation}
    However, from \Eref{eq: Malus} we can see that the polarized image $\mathbf{I}_\alpha$ can be regarded as a re-exposed version of the unpolarized image $\mathbf{I}$ with spatially-variant and arbitrary exposures (\ie, $\mathbf{I}_\alpha = \mathbf{I} \cdot \mathbf{M}$ where $\mathbf{M} = \frac{1}{2}(1 - \mathbf{p} \cdot \cos(2(\alpha-\bm{\theta}))) \in [0,1]$ is a scene-dependent exposure mask) \citep{wu2020hdr, zhou2023polarizationHDR}, which poses a major challenge to efficiently extract high-level features (\eg, semantic and contextual information) from $\mathbf{B}_{\alpha_{1,2,3,4}}$ for deblurring. Particularly, we have observed that the polarized images usually have similar semantic and contextual information to the corresponding unpolarized image. To verify it, we design another simulation experiment on our training set to quantitatively measure the gap between $\mathbf{I}$ and $\mathbf{I}_{\alpha_{1,2,3,4}}$, by using the contrastive language-image pre-training (CLIP) model \citep{radford2021learning} to obtain their encoded semantic feature maps and compute the average value of MSE between them (which we defined as the semantic distance) \citep{chefer2022image}. Results are shown in \fref{fig: Verification} (right), we can see that the semantic distance between $\mathbf{I}_{\alpha_{1,2,3,4}}$ and $\mathbf{I}$ could be very small (even nearly one order of magnitude smaller than that between $\mathbf{I}_{\alpha_{1,2,3,4}}$ and $\mathbf{B}_{\alpha_{1,2,3,4}}$). Therefore, we choose to transfer the semantic and contextual information encoded in the estimated unpolarized image $\mathbf{I}_{\text{guide}}$ to the target polarized images $\mathbf{I}_{\alpha_{1,2,3,4}}$.  

    \textbf{Layer details.} Similar to $g_1$, we adopt another backbone network $\mathcal{B}_2$ to learn the residual between $\mathbf{B}_{\alpha_{1,2,3,4}}$ and $\mathbf{I}_{\alpha_{1,2,3,4}}$. Since we want semantic information to flow from $\mathbf{I}_{\text{guide}}$ to $\mathbf{B}_{\alpha_{1,2,3,4}}$ for guidance, instead of directly feeding the concatenated features of $\mathbf{B}_{\alpha_{1,2,3,4}}$ and $\mathbf{I}_{\text{guide}}$ into $\mathcal{B}_2$, we propose to use multiscale context transferring encoder (MCTE) to substitute the encoder inside $\mathcal{B}_2$ for processing them. MCTE adopts several convolution layers to perform context transferring in each scale, and then uses residual blocks \citep{he2016deep} to refine the features for enriching details. Specifically, we adopt another feature downsampling block $\mathcal{D}_2$ to downsample the features of $\mathbf{I}_{\text{guide}}$ extracted by a feature extraction head $\mathcal{H}_u$, and then feed them into MCTE along with the features of $\mathbf{B}_{\alpha_{1,2,3,4}}$ extracted by another feature extraction head $\mathcal{H}_p$. Besides, plugging the values of $\alpha_{1,2,3,4}$ into \Eref{eq: Malus}, we can see that $\mathbf{I}=\mathbf{I}_{\alpha_1} + \mathbf{I}_{\alpha_3}=\mathbf{I}_{\alpha_2}+\mathbf{I}_{\alpha_4}$. So, we propose a polarization-based grouping (PG) strategy, by dividing $\mathbf{B}_{\alpha_{1,2,3,4}}$ into two groups ($\mathbf{B}_{\alpha_{1,3}}$ and $\mathbf{B}_{\alpha_{2,4}}$) and adopting two sets of network modules with shared parameters (including $\mathcal{H}_p$ and $\mathcal{B}_2$) to deal with them respectively. In this way, the problem could be further decomposed into two guided image decomposition problems: separating two image layers $\mathbf{I}_{\alpha_{1,3}}$ (or $\mathbf{I}_{\alpha_{2,4}}$) from a ``mixture image'' $\mathbf{I}_{\text{guide}}$ under the guidance of their blurry counterparts $\mathbf{B}_{\alpha_{1,3}}$ (or $\mathbf{B}_{\alpha_{2,4}}$), which could be more robust than jointly processing $\mathbf{B}_{\alpha_{1,2,3,4}}$ in a direct manner.

\section{Implementation details}
    \subsection{Loss function}
    \label{subsec: Loss}
    The total loss function $L$ can be written as
    \begin{equation}
        \begin{split}
        &L(\mathbf{I}_{\text{guide}}, \mathbf{I}_{\alpha_{1,2,3,4}}, \mathbf{S}_{0,1,2})\\ 
        &= \lambda_1 L_c(\mathbf{I}_{\text{guide}}) + \lambda_2 \sum_{i=1}^{4} L_c(\mathbf{I}_{\alpha_i}) + \lambda_3 L_s(\mathbf{S}_0, \mathbf{S}_1, \mathbf{S}_2),
        \end{split}
    \end{equation}
    where $\lambda_{1,2,3}$ are set to be 0.5, 1, and 1 respectively, $L_c$ (content loss) and $L_s$ (Stokes loss) denote two kinds of basic loss functions used during the training process respectively. We will detail each of them in the following.
    
    \textbf{Content loss.} The content loss $L_c$ aims to bridge the gap between the restored image $\mathbf{I}_\forall$ (which can be any one of $\mathbf{I}_{\text{guide}}$ and $\mathbf{I}_{\alpha_{1,2,3,4}}$) and its corresponding ground truth for sharper image content, which is defined as
    \begin{equation}
        \begin{split}
        L_c(\mathbf{I}_\forall) 
        &= \lambda_{c_1} L_1(\mathbf{I}_\forall) + \lambda_{c_2} L_2(\mathbf{I}_\forall)\\
        &+ \lambda_{c_3} L_\text{perc}(\mathbf{I}_\forall) + \lambda_{c_4} L_\text{edge}(\mathbf{I}_\forall),
        \end{split}
    \end{equation}
    where $L_1$, $L_2$, $L_\text{perc}$, and $L_\text{edge}$ denote the $\ell_1$, $\ell_2$, perceptual, and edge loss respectively, and $\lambda_{c_{1,2,3,4}}$ are set to be 10, 100, 0.1, and 10 respectively. The perceptual loss $L_\text{perc}$ is defined as the $\ell_2$ loss computed using the feature maps extracted by $VGG_{3,3}$ convolution layer of VGG-19 network \citep{simonyan2014very} pretrained on ImageNet \citep{russakovsky2015imagenet}, and the edge loss $L_\text{edge}$ is defined as the $\ell_2$ loss computed using the edges extracted by Laplace kernels. 
    
    \textbf{Stokes loss.} The Stokes loss $L_s$ aims to bridge the gap between the restored Stokes parameters $\mathbf{S}_{0,1,2}$ (computed from the restored polarized images $\mathbf{I}_{\alpha_{1,2,3,4}}$ using \Eref{eq: StokesParameters}) and their corresponding ground truth for more accurate polarization constraints, which is defined as
    \begin{equation}
        \label{eq: StokesLoss}
        \begin{split}
        L_s(\mathbf{S}_0, \mathbf{S}_1, \mathbf{S}_2) 
        &= \lambda_{s_1} L_2(\mathbf{S}_0) + \lambda_{s_2} L_2(\mathbf{S}_1)\\
        &+ \lambda_{s_3} L_2(\mathbf{S}_2) + \lambda_{s_4} L_2(\frac{\mathbf{S}_2}{\mathbf{S}_1}),
        \end{split}
    \end{equation}
    where $\lambda_{s_{1,2,3,4}}$ are set to be 20, 500, 500, and 500 respectively. Note that the fourth term of $L_s$ is adopted to optimize the ratio between $\mathbf{S}_2$ and $\mathbf{S}_1$ since the calculation of the AoLP depends on it (see \Eref{eq: AoP}).
    
    \subsection{Dataset preparation}
    \label{subsec: Dataset}
    There is no public dataset for such a polarized image deblurring task, and it is difficult to obtain pairwise blurry and sharp polarized images at a large scale. Besides, existing deblurring benchmark datasets (\eg, GOPRO \citep{nah2017deep}, HIDE \citep{shen2019human}, and RealBlur \citep{rim_2020_ECCV}) do not contain any polarization information, which cannot be used to generate polarized images. Therefore, we propose to generate a synthetic dataset for training our network. The generating process can be described as the following steps:
    \begin{itemize}
        \item[(1)] Collect polarized images from two public polarized low-light image enhancement datasets (LLCP \citep{xu2022colorpolarnet} and PLIE \citep{zhou2023polarization}) to serve as the ground-truth sharp images.
        \item[(2)] Randomly generate camera shake trajectories using the algorithm proposed by \cite{boracchi2012modeling}.
        \item[(3)] Adopt the spatially-variant blur model proposed by \cite{whyte2012non} to make the trajectories pixel-wise.
        \item[(4)] Move the corresponding ground-truth sharp images along the pixel-wise trajectories to get multiple sharp latent frames.
        \item[(5)] Averaging those sharp latent frames to synthesize the blurry images.
        \item[(6)] Add noise to the blurry images (as done in other works \citep{hu2020iplnet, xu2022colorpolarnet, zhou2023polarization}) to better simulate the real situation. 
    \end{itemize}
    
    The LLCP \citep{xu2022colorpolarnet} and PLIE \citep{zhou2023polarization} datasets provide 430 different scenes in total, and we randomly split them into two parts that contain 400 and 30 scenes for making the training and test sets respectively. For each scene in the training (test) set, we randomly generate 20 (10) different camera shake trajectories and further perform data augmentation (\eg, flipping and rotating) so that the training (test) set contains 8000 (300) different images finally. Note that data augmentation is only applied when generating the training set. The images are resized and randomly cropped to $256 \times 256$ ($512 \times 512$) pixels in the training (test) set.

    \begin{table*}[t]
        \caption{Quantitative comparisons on synthetic data among our method, IPLNet \citep{hu2020iplnet}, ColorPolarNet \citep{xu2022colorpolarnet}, PLIE \citep{zhou2023polarization}, MIMO \citep{cho2021rethinking}, XYDeblur \citep{ji2022xydeblur}, STDAN \citep{zhang2022spatio}, and MISCFilter \citep{liu2024motion}. \textbf{Bold} font indicates the best performance.}
        \label{tab: SyntheticData}
        \centering
        \begin{tabular}{lcccccc}
            \toprule
            & PSNR-$\mathbf{p}$ & SSIM-$\mathbf{p}$ & PSNR-$\bm{\theta}$ & SSIM-$\bm{\theta}$ & PSNR-$\mathbf{I}$ & SSIM-$\mathbf{I}$ \\
            \midrule
            Ours                                      & \textbf{29.12} & \textbf{0.774} & \textbf{17.77} & \textbf{0.330} & \textbf{29.67} & \textbf{0.868}\\
            IPLNet \citep{hu2020iplnet}	              & 19.98          & 0.702          & 17.01          & 0.267          & 26.08          & 0.822\\
            ColorPolarNet \citep{xu2022colorpolarnet} & 28.41          & 0.740          & 16.99          & 0.279          & 25.09          & 0.812\\
            PLIE \citep{zhou2023polarization}         & 26.32          & 0.744          & 17.24          & 0.269          & 28.40          & 0.853\\
            MIMO \citep{cho2021rethinking}            & 26.61          & 0.730          & 16.51          & 0.280          & 25.92          & 0.830\\
            XYDeblur \citep{ji2022xydeblur}           & 22.74          & 0.698          & 15.83          & 0.272          & 26.68          & 0.850\\
	    STDAN \citep{zhang2022spatio}             & 28.28          & 0.756          & 17.06          & 0.309          & 27.49          & 0.862\\
            MISCFilter \citep{liu2024motion}          & 28.75          & 0.763          & 17.16          & 0.311          & 28.45          & 0.865\\
            \bottomrule
        \end{tabular}
    \end{table*}
    
    \begin{figure*}[t]
        \centering
        \includegraphics[width=1.0\linewidth]{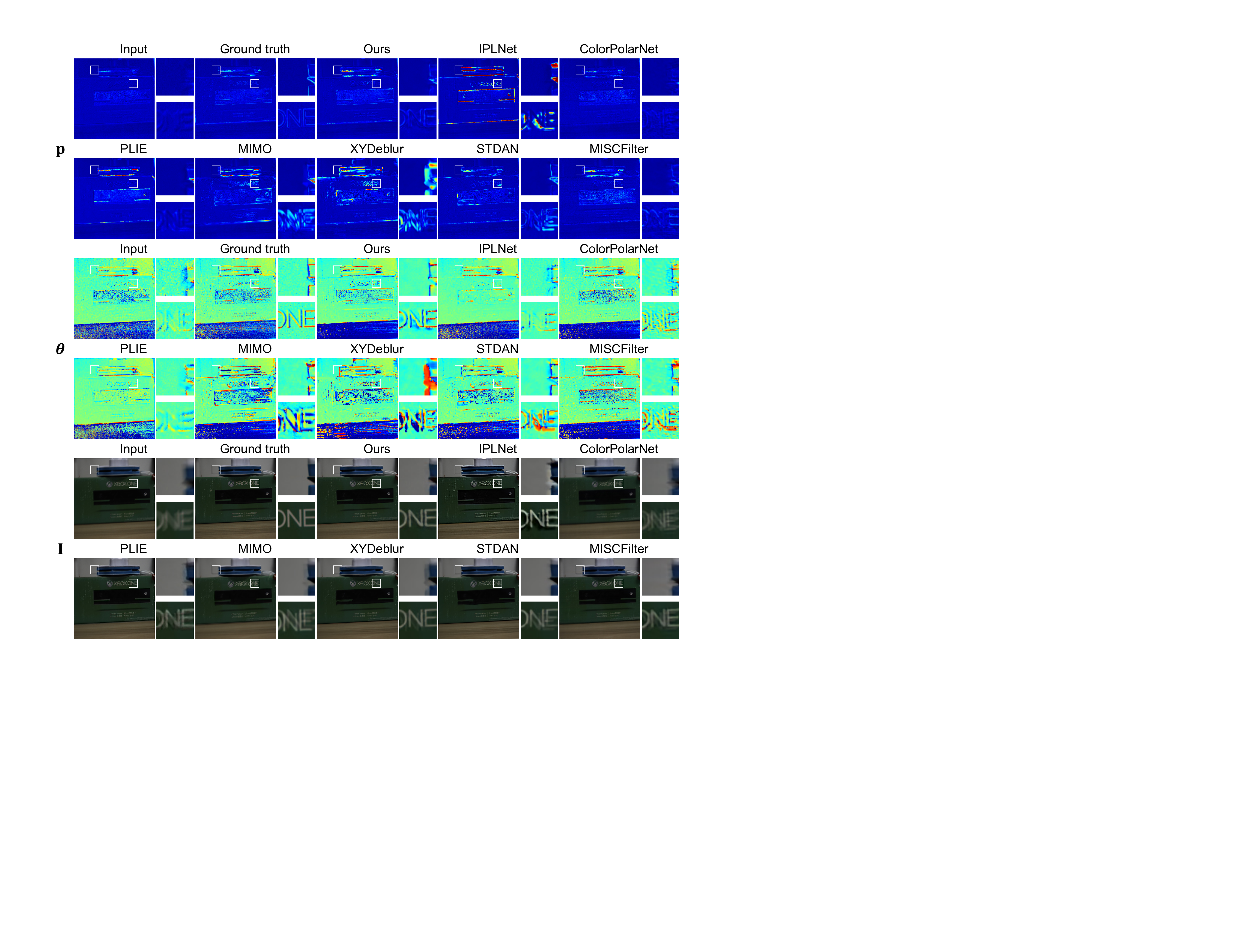}
        \caption{Qualitative comparisons on synthetic data among our method, IPLNet \citep{hu2020iplnet}, ColorPolarNet \citep{xu2022colorpolarnet}, PLIE \citep{zhou2023polarization}, MIMO \citep{cho2021rethinking}, XYDeblur \citep{ji2022xydeblur}, STDAN \citep{zhang2022spatio}, and MISCFilter \citep{liu2024motion}. The DoLP $\mathbf{p}$ and AoLP $\bm{\theta}$ are visualized using color maps after normalizing and averaging the RGB channels (as done in \citep{hu2020iplnet, zhou2023polarization}). Please zoom-in for better details.}
        \label{fig: SyntheticData}
    \end{figure*}
    
    \subsection{Training strategy}
    \label{subsec: Training}
    We implement our method using PyTorch with 4 NVIDIA 1080Ti GPUs and apply a two-stage training strategy. First, to ensure a stable initialization of the training process, we train the unpolarized image estimator and the polarized image reconstructor independently for 600 and 300 epochs respectively with a learning rate of 0.002. Then, we finetune the entire network in an end-to-end manner for another 100 epochs with a learning rate of 0.001. 
    
    For optimization, we use Adam optimizer \citep{kingma2014adam} with $\beta_1=0.5$, $\beta_2=0.999$. The network parameters are initialized with Xavier initialization \citep{glorot2010understanding}. We add an instance normalization \citep{ulyanov2016instance} layer and a \texttt{ReLU} activation function after each convolution layer. 

    \begin{table*}[t]
        \caption{Quantitative comparisons on real data among our method, IPLNet \citep{hu2020iplnet}, ColorPolarNet \citep{xu2022colorpolarnet}, PLIE \citep{zhou2023polarization}, MIMO \citep{cho2021rethinking}, XYDeblur \citep{ji2022xydeblur}, STDAN \citep{zhang2022spatio}, and MISCFilter \citep{liu2024motion}. \textbf{Bold} font indicates the best performance.}
        \label{tab: RealData}
        \centering
        \begin{tabular}{lcccccc}
            \toprule
            & CLIP-IQA-$\mathbf{p}$ & CLIP-IQA-$\bm{\theta}$ & CLIP-IQA-$\mathbf{I}$\\
            \midrule
            Ours                                      & \textbf{0.506} & \textbf{0.677} & \textbf{0.439}\\
            IPLNet \citep{hu2020iplnet}               & 0.503          & 0.628          & 0.428\\
            ColorPolarNet \citep{xu2022colorpolarnet} & 0.479          & 0.592          & 0.366\\
            PLIE \citep{zhou2023polarization}         & 0.484          & 0.589          & 0.274\\
            MIMO \citep{cho2021rethinking}            & 0.500          & 0.667          & 0.343\\
            XYDeblur \citep{ji2022xydeblur}           & 0.504          & 0.624          & 0.359\\
	    STDAN \citep{zhang2022spatio}             & 0.479          & 0.593          & 0.366\\
     	MISCFilter \citep{liu2024motion}          & 0.433          & 0.582          & 0.361\\
            \bottomrule
        \end{tabular}
    \end{table*}

\section{Experiments}    
    \subsection{Evaluation on synthetic data}
    Since there is no existing approach specially designed for deblurring polarized images, we choose to compare our method to two groups of methods that solve the closest problems. The first group is originally designed for handling the polarized image enhancement tasks, including IPLNet \citep{hu2020iplnet}, ColorPolarNet \citep{xu2022colorpolarnet}, and PLIE \citep{zhou2023polarization}. The second group is originally designed for handling the unpolarized image/video deblurring tasks, including MIMO \citep{cho2021rethinking}, XYDeblur \citep{ji2022xydeblur}, STDAN \citep{zhang2022spatio}, and MISCFilter \citep{liu2024motion}. All methods in the first group can process the polarized images simultaneously in a direct manner. In the second group, STDAN \citep{zhang2022spatio} can process the polarized images simultaneously in an indirect manner since it is designed for deblurring continuous video frames, while MIMO \citep{cho2021rethinking}, XYDeblur \citep{ji2022xydeblur}, and MISCFilter \citep{liu2024motion} can only process the polarized images in a frame-by-frame manner since it is designed for deblurring a single image. Note that since STDAN \citep{zhang2022spatio} reconstructs the middle frame by referencing 5 neighboring frames, when compared to it we set its input as a video whose frames are stacked by the blurry polarized images in the order of ``$[\mathbf{B}_{\alpha_3},\mathbf{B}_{\alpha_4},\mathbf{B}_{\alpha_1},\mathbf{B}_{\alpha_2},\mathbf{B}_{\alpha_3},\mathbf{B}_{\alpha_4},\mathbf{B}_{\alpha_1},\mathbf{B}_{\alpha_2}]$'' to ensure that the output frames are in the order of ``$[\mathbf{B}_{\alpha_1},\mathbf{B}_{\alpha_2},\mathbf{B}_{\alpha_3},\mathbf{B}_{\alpha_4}]$''. 

    All the compared methods are trained on our dataset (either finetuned or retrained for adaptation to our dataset) under the supervision of the polarized images $\mathbf{I}_{\alpha_{1,2,3,4}}$ (the very same outputs of our method). In practice, we choose to finetune (\ie, train from pre-trained weights) the methods designed for deblurring unpolarized images \citep{cho2021rethinking, ji2022xydeblur, liu2024motion, zhang2022spatio} and retrain (\ie, train from scratch) the methods designed for enhancing polarized images \citep{hu2020iplnet, xu2022colorpolarnet, zhou2023polarization}. Such a training strategy for the compared methods could be reasonable and often adopted by other methods \citep{hu2020iplnet, xu2022colorpolarnet, zhou2023polarization}, since finetuning is commonly used when adapting a model to a specific task or domain that is closely related to the original task the model was trained on (\eg, deblurring unpolarized images \vs deblurring polarized images), while retraining is typically used when the task is different from the original tasks that pre-trained models cannot provide a good starting point (\eg, enhancing low-light polarized images \vs deblurring polarized images). Besides, for a fair comparison, we apply our loss function to the methods designed for enhancing polarized images \citep{hu2020iplnet, xu2022colorpolarnet, zhou2023polarization}, while preserving the original loss functions for the methods designed for deblurring unpolarized images \citep{cho2021rethinking, ji2022xydeblur, liu2024motion, zhang2022spatio} during training. This is because our loss function includes a Stokes loss term (\Eref{eq: StokesLoss}), which requires multiple polarized images to compute the polarization-related parameters, while multiple polarized images cannot be obtained within a single forward pass for the methods designed for deblurring unpolarized images. Additionally, forcibly modifying the forward/backward propagation logic of these methods to accommodate our loss function may degrade their performance rather than improve it.

    \begin{figure*}[t]
        \centering
        \includegraphics[width=1.0\linewidth]{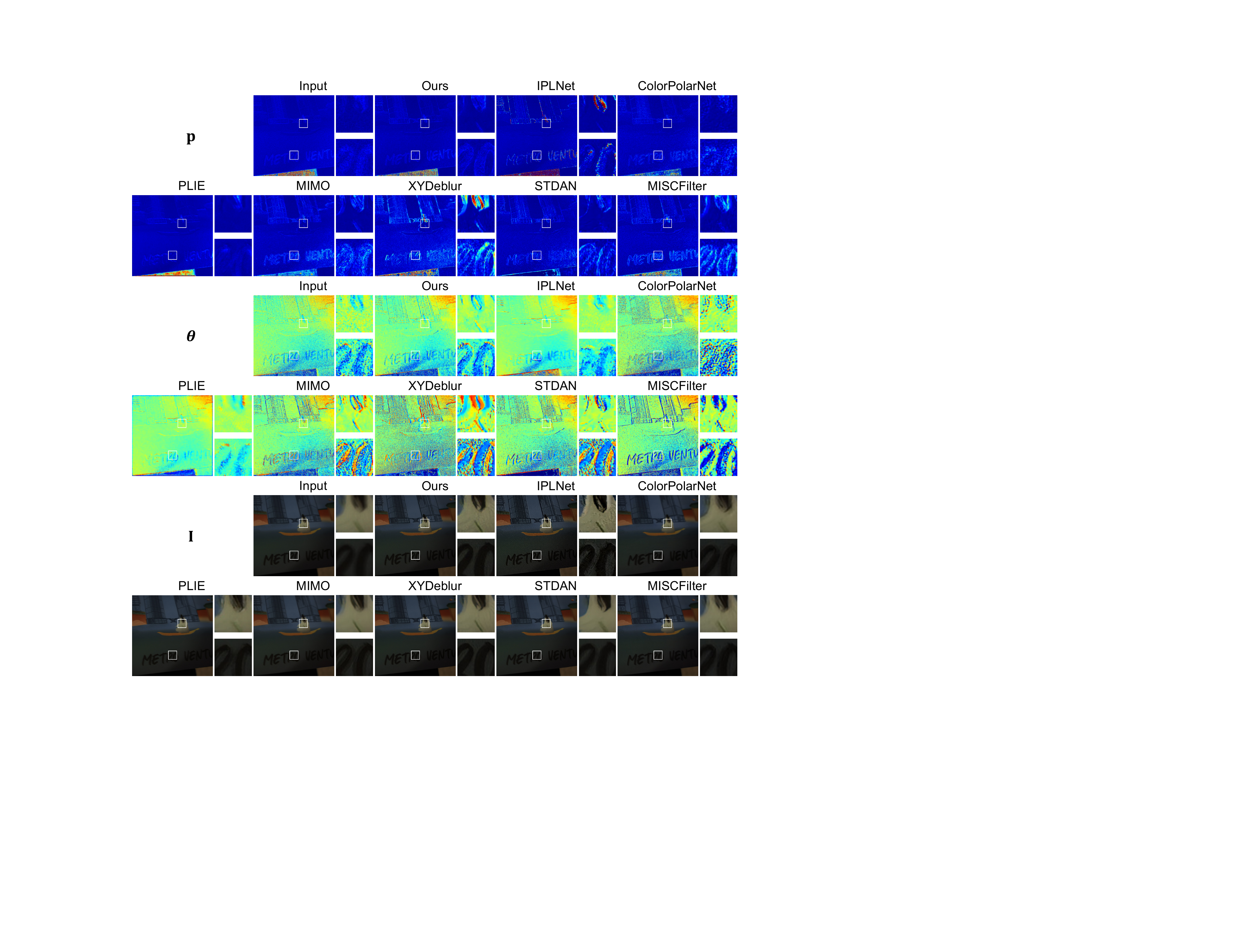}
        \caption{Qualitative comparisons on real data among our method, IPLNet \citep{hu2020iplnet}, ColorPolarNet \citep{xu2022colorpolarnet}, PLIE \citep{zhou2023polarization}, MIMO \citep{cho2021rethinking}, XYDeblur \citep{ji2022xydeblur}, STDAN \citep{zhang2022spatio}, and MISCFilter \citep{liu2024motion}. The DoLP $\mathbf{p}$ and AoLP $\bm{\theta}$ are visualized using color maps after normalizing and averaging the RGB channels (as done in \citep{hu2020iplnet, zhou2023polarization}). Please zoom-in for better details.}
        \label{fig: RealData}
    \end{figure*}

    To evaluate the results quantitatively, we choose to adopt two frequently-used metrics including PSNR and SSIM. Results are shown in \Tref{tab: SyntheticData}. Our model consistently outperforms the compared methods on all metrics. Visual quality comparisons are shown in \fref{fig: SyntheticData}\footnote{Additional results can be found in the supplementary material.}. Note that we not only evaluate the accuracy of the restored DoLP $\mathbf{p}$ and AoLP $\bm{\theta}$, but also evaluate the quality of the restored unpolarized image $\mathbf{I}$. As for $\mathbf{p}$ and $\bm{\theta}$, our method achieves better performance than other methods, because we explicitly consider preserving the polarization constraints. Taking the second box of $\mathbf{p}$ as an example, the compared methods either suffer from severe artifacts (\eg, IPLNet \citep{hu2020iplnet}, MIMO \citep{cho2021rethinking}, XYDeblur \citep{ji2022xydeblur}, and STDAN \citep{zhang2022spatio}) or tend to generate over-smooth edges (\eg, ColorPolarNet \citep{xu2022colorpolarnet}, PLIE \citep{zhou2023polarization}, and MISCFilter \citep{liu2024motion}). This is because the methods designed for deblurring can only process the polarized images in a polarization-unaware manner, while the methods designed for enhancing polarized images do not take the unique properties of motion blur into consideration. As for $\mathbf{I}$, our results resemble the ground truth more closely with less distortion. This is because our method can properly deal with the degraded low-level features and less distinctive high-level features.

    \begin{table*}[t]
        \caption{Quantitative evaluation results of ablation study on synthetic data. \textbf{Bold} font indicates the best performance.}
        \label{tab: AblationStudy}
        \centering
        \begin{tabular}{lcccccc}
            \toprule
            & PSNR-$\mathbf{p}$ & SSIM-$\mathbf{p}$ & PSNR-$\bm{\theta}$ & SSIM-$\bm{\theta}$ & PSNR-$\mathbf{I}$ & SSIM-$\mathbf{I}$ \\
            \midrule
            W/o $\mathbf{I}_{\text{guide}}$ & 27.12 & 0.730 & 17.14 & 0.241 & 28.36 & 0.856\\
            W/o $\mathbf{S}^*_{1,2}$ & 28.41 & 0.772 & 17.67 & 0.329 & 29.16 & 0.863\\
            $\mathbf{G}$ instead of $\mathbf{S}^*_{1,2}$ & 28.43 & 0.768 & 17.70 & 0.329 & 29.19 & 0.864\\
            W/o SGSC & 29.11 & 0.773 & 17.71 & 0.327 & 28.99 & 0.862\\
            W/o MCTE & 28.63 & 0.760 & 17.17 & 0.267 & 29.38 & 0.866\\
            W/o PG & 29.10 & 0.773 & 17.52 & 0.293 & 29.50 & 0.866\\
            W/o $L_s$ & 29.03 & 0.769 & 17.48 & 0.314 & 29.62 & 0.867\\
            Our complete model & \textbf{29.12} & \textbf{0.774} & \textbf{17.77} & \textbf{0.330} & \textbf{29.67} & \textbf{0.868}\\
            \bottomrule
        \end{tabular}
    \end{table*}

    \begin{figure*}[t]
        \centering
        \includegraphics[width=1.0\linewidth]{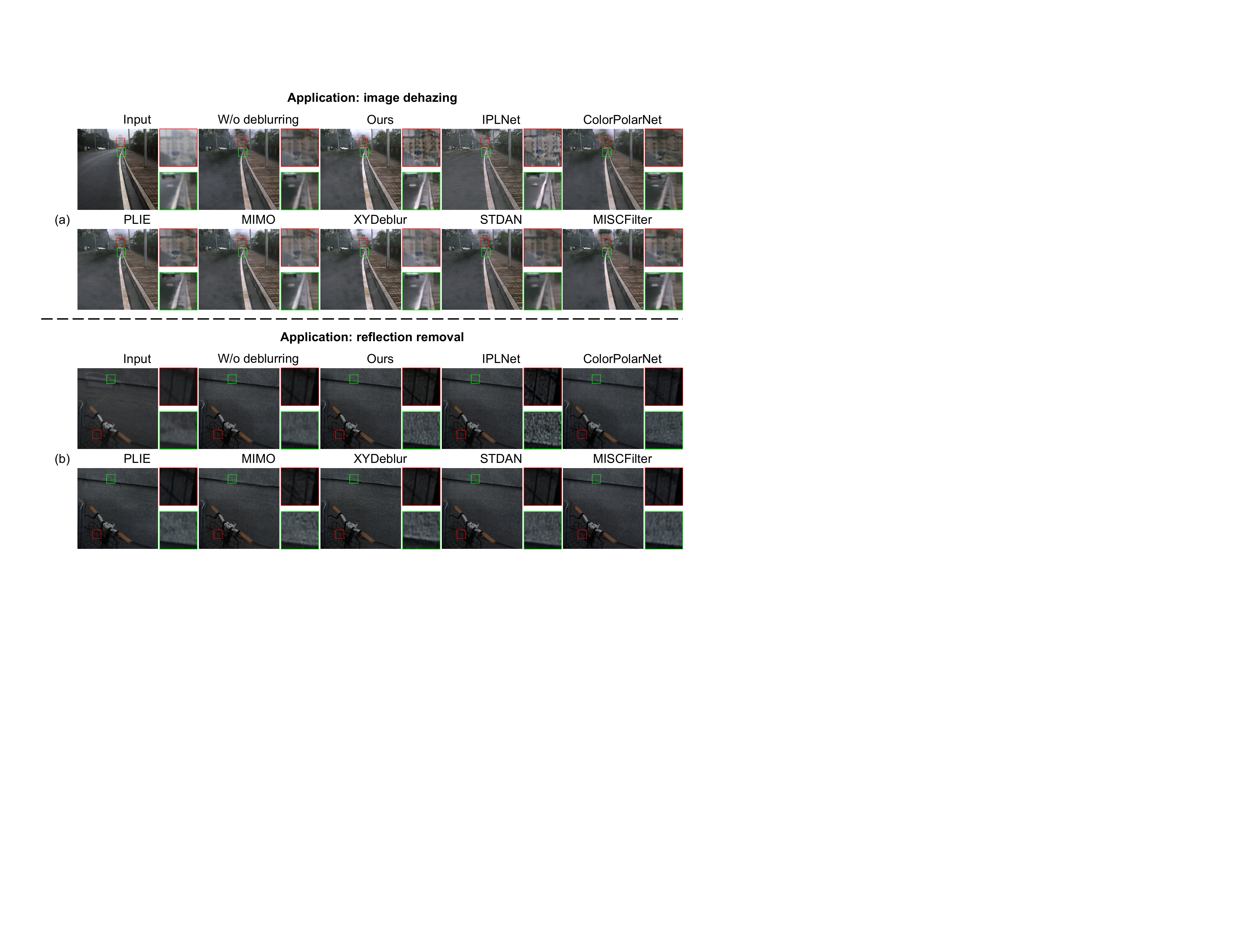}
        \caption{Results of polarization-based vision applications (including (a) image dehazing \citep{zhou2021learning} and (b) reflection removal \citep{lyu2019reflection, lyu2022physics}) without deblurring and deblurred by our method and the compared methods (IPLNet \citep{hu2020iplnet}, ColorPolarNet \citep{xu2022colorpolarnet}, PLIE \citep{zhou2023polarization}, MIMO \citep{cho2021rethinking}, XYDeblur \citep{ji2022xydeblur}, STDAN \citep{zhang2022spatio}, and MISCFilter \citep{liu2024motion}). Please zoom-in for better details.}
        \label{fig: Application}
    \end{figure*}
    
    \subsection{Evaluation on real data}
    To demonstrate the strong generalization capability of our method on complex and diverse real-world data, we capture a real-world dataset using a Lucid Vision Phoenix polarization camera. The dataset comprises 40 different scenes, which exhibits a range of conditions (\eg, different environments, varying blur patterns, \etc). Note that the dataset does not provide the ground truth for reference.

    To evaluate the results quantitatively, we choose to adopt a non-reference metric, CLIP-IQA \citep{wang2023exploring} (contrastive language-image pretraining for image quality assessment). The results, presented in \Tref{tab: RealData}, indicate that our model outperforms the compared methods, showcasing superior performance across the dataset. Visual quality comparisons are shown in \fref{fig: RealData}\footnote{Additional results can be found in the supplementary material.}. From the results we can see that our method suffer less from artifacts. Taking the first box of $\mathbf{p}$ as an example, the compared methods tend to produce inaccurate results (shown as abnormal colors). Taking the first box of $\mathbf{I}$ as another example, we can see that IPLNet \citep{hu2020iplnet} brings about ringing artifacts with higher noise level, while other compared methods fail to restore sharp image content.

    \begin{table*}[t]
        \caption{Computational complexity analysis on synthetic data among our method, IPLNet \citep{hu2020iplnet}, ColorPolarNet \citep{xu2022colorpolarnet}, PLIE \citep{zhou2023polarization}, MIMO \citep{cho2021rethinking}, XYDeblur \citep{ji2022xydeblur}, STDAN \citep{zhang2022spatio}, and MISCFilter \citep{liu2024motion}. Here, the size of the input images is $512 \times 512$ pixels (3 color channels), and the inference time is the time taken to reconstruct all 300 scenes.}
        \label{tab: ComputationalComplexity}
        \centering
        \begin{tabular}{lcccccc}
            \toprule
            & Number of parameters & FLOPs & Memory & Inference time\\
            \midrule
            Ours                                      & 3.14 M  & 360.20 G  & 2.39 G & 4.36 s\\
            IPLNet \citep{hu2020iplnet}               & 1.80 M  & 944.97 G  & 1.89 G & 3.57 s\\
            ColorPolarNet \citep{xu2022colorpolarnet} & 1.03 M  & 943.94 G  & 1.51 G & 1.90 s\\
            PLIE \citep{zhou2023polarization}         & 1.50 M  & 453.94 G  & 4.28 G & 4.84 s\\
            MIMO \citep{cho2021rethinking}            & 16.11 M & 1207.58 G & 2.46 G & 13.56 s\\
            XYDeblur \citep{ji2022xydeblur}           & 4.92 M  & 550.40 G  & 2.18 G & 12.19 s\\
	      STDAN \citep{zhang2022spatio}             & 13.84 M & 3766.04 G & 5.94 G & 129.73 s\\
     	MISCFilter \citep{liu2024motion}          & 15.99 M & 993.29 G  & 2.82 G & 60.92 s\\
            \bottomrule
        \end{tabular}
    \end{table*}
   
    \subsection{Ablation study}
    To verify the validity of each design choice, we conduct a series of ablation studies, which can be expressed as
    \begin{itemize}
        \item[(1)] W/o $\mathbf{I}_{\text{guide}}$: deblurring $\mathbf{B}_{\alpha_{1,2,3,4}}$ directly without estimating $\mathbf{I}_{\text{guide}}$ first (to demonstrate the contribution of our two-stage deblurring pipeline).
        \item[(2)] W/o $\mathbf{S}^*_{1,2}$: removing $\mathbf{S}^*_{1,2}$ from $g_1$ (to verify the necessity of using $\mathbf{S}^*_{1,2}$ as priors).
        \item[(3)] $\mathbf{G}$ instead of $\mathbf{S}^*_{1,2}$: 
        substituting the Stokes parameters $\mathbf{S}^*_{1,2}$ with the gradient distribution $\mathbf{G}$ (to show the advantage of using $\mathbf{S}^*_{1,2}$ over using $\mathbf{G}$).
        \item[(4)] W/o SGSC: directly feeding the fused Stokes features $\mathbf{F}$ into the encoder of $\mathcal{B}_1$ (to show the importance of SGSC used in $g_1$).
        \item[(5)] W/o MCTE: directly feeding the concatenated features of $\mathbf{B}_{\alpha_{1,2,3,4}}$ and $\mathbf{I}_{\text{guide}}$ into $\mathcal{B}_2$ (to show the importance of MCTE used in $g_2$).
        \item[(6)] W/o PG: using a single backbone network to process $\mathbf{B}_{\alpha_{1,2,3,4}}$ jointly (to validate the effectiveness of adopting the PG strategy in $g_2$).
        \item[(7)] W/o $L_s$: removing the Stokes loss (to show the benefit of using such kind of loss in our setting).
    \end{itemize}
    Quantitative comparisons are shown in \Tref{tab: AblationStudy}. These results demonstrate that our complete model achieves the optimal performance with these specific design choices.

    \subsection{Applications}
    To show the benefits of deblurring polarized images, we choose two different polarization-based vision applications including image dehazing \citep{zhou2021learning} and reflection removal \citep{lyu2019reflection, lyu2022physics} as the validation methods respectively, and show that deblurring can improve their performance. Comparisons of the haze-removed and reflection-removed unpolarized images are shown in \fref{fig: Application} (a) and (b) respectively\footnote{Additional results can be found in the supplementary material.}, which demonstrate that our method brings superior performance gain than the compared methods. Taking the red box in \fref{fig: Application} (a) as an example, we can see that the haze cannot be removed robustly without deblurring, and the results deblurred by the compared methods are still hazy, while the result deblurred by our method has higher clarity. Taking the green box in \fref{fig: Application} (b) as another example, we can see that the result deblurred by our method has sharp textures with less reflection contamination, while IPLNet \citep{hu2020iplnet} alters the contrast of the input image excessively and generates fake textures that originally did not exist, and other compared methods fail to produce sharp textures.

    \subsection{Computational complexity analysis}
    To evaluate the computational complexity, we compare the number of parameters, FLOPs, memory, and inference time of different methods on our synthetic test dataset using a single NVIDIA 4090 GPU, as shown in \Tref{tab: ComputationalComplexity}. Here, the size of the input images is $512 \times 512$ pixels (3 color channels), and the inference time is the time taken to reconstruct all 300 scenes. Besides, since the methods designed for handling the unpolarized image deblurring tasks (MIMO \citep{cho2021rethinking}, XYDeblur \citep{ji2022xydeblur}, and MISCFilter \citep{liu2024motion}) cannot simultaneously process four polarized images, the total time for processing four polarized images is taken as the inference time. From the results we can see that the computational complexity of our method is not very high, and it can realize real-time deblurring of polarized images.

\section{Conclusion}
    \label{sec: Conclusion}
    We present a learning-based solution to deblur multiple polarized images for acquiring the DoLP and AoLP accurately. Specifically, we propose a polarized image deblurring pipeline to solve the problem in a polarization-aware manner by decomposing it into two less-ill-posed sub-problems, and build a two-stage neural network tailored to the pipeline to handle the two sub-problems respectively. Experimental results not only show that our method achieves state-of-the-art performance on both synthetic and real-world images, but also demonstrate that using our method to deblur polarized images can improve the performance of polarization-based vision applications such as reflection removal and image dehazing.
    
    \textbf{Limitations.} Since our method is specially designed for deblurring multiple polarized images captured by a polarization camera in a snapshot, it cannot be used to deblur videos since it does not explicitly consider the temporal information encoded in consecutive frames. Besides, since the formation of motion blur could be different from other types of degeneration (such as haze, rain, and noise), our method, which is based on the observations about motion blur, may not generalize to other low-level tasks.

    \textbf{Discussions.} In the future, we aim to extend our work to address the polarized video deblurring problem by developing new modules that can better explore and utilize temporal connections. Additionally, we plan to generalize our approach to handle other types of image degradation by integrating comprehensive observations about polarized images, ultimately working towards a unified framework for enhancing polarized images.

\section*{Acknowledgement} This work was supported in part by National Natural Science Foundation of China under Grant No. 62136001, 62088102, 62276007, Beijing Municipal Science \& Technology Commission, Administrative Commission of Zhongguancun Science Park (Grant No. Z241100003524012), and JST-Mirai Program Grant Number JPMJM123G1.

{\small
\bibliographystyle{spbasic}
\bibliography{egbib}
}

\end{document}